# Enhancing English Writing Proficiency in China's Polytechnic Students: An In-Depth Literature Review on the Application of the Input Hypothesis

ZhouWei

Jeonbuk National University

## Abstract

Having good English writing skills is extremely important for students in polytechnic institutions. However, a lot of students in technical schools have difficulties in reaching high levels of skill. The Input Hypothesis, created by Stephen Krashen, suggests that people learn languages well when they receive information that's a little harder than what they already know but still understandable. This research paper wants to study how the Input Hypothesis can help polytechnic students improve their English writing skills. The study will include real-life observations and experiments from the previous research. We will look at data from polytechnic students who are receiving special writing instruction to see if the Input Hypothesis actually helps improve their writing skills. The paper can better inform polytechnic students, faculty members, and support staff and even members of the larger community about the attributions, the processes, and the possible outcomes of second language development for polytechnic students.

**Keywords: English writing skills, Polytechnic students, Input hypothesis, Comprehensible input**

## Introduction

Having good English writing skills is very important for polytechnic students to do well in their studies and have better job opportunities in the future. According to the former study, it showed that English is a global language that substantially affects how the sector views its employees and how individuals communicate. Excellent English communication abilities lead to a more favorable impression of an individual in a professional setting (Siregar et al., 2022). Teaching English as a foreign language (EFL) in Vocational Higher Education has a goal not only for introducing English at the academic level, but also of gaining students' interest in placing the four notions of skills (listening, speaking, reading, and writing) as a part of individual practical use in industrial practice (Maldin, 2022). This becomes a massive challenge for those teachers in EFL classrooms to build interest and unleash students' motivation for making use of English in daily life activities. It was reported from a survey of entrepreneurs and college graduates that "English skills, especially English reading, writing and translating, are frequently used in works. Most graduates from Higher Vocational Colleges pay more attention to the English material related to their majors, and they use English mostly in reading product descriptions, operation manuals, and equipment specifications and write a report (Xia, 2019). A lot of students find it difficult to become good at writing in English. This paper tries to find out if the Input Hypothesis can help polytechnic students improve their English writing skills. The Input Hypothesis theory, made by Stephen Krashen, suggests that people learn languages best when they are given understandable information that is a little bit harder than what they currently understand. The Input Hypothesis (IH) assumes that we acquire language by understanding messages. More precisely, comprehensible input is the essential environmental ingredient--a richly specified internal language acquisition device also makes a significant contribution to language acquisition. I argue that the best hypothesis is that competence in spelling and vocabulary is most efficiently attained by comprehensible input in the form of reading, a position argued by several others (Krashen, 2023). To see how this theory affects students in technical schools, this research paper presents a study that looks at specific writing instruction. This review wants to give a complete summary of past research on this topic and how it can help polytechnic students improve their English writing skills.

## 1. Theoretical Foundations

**1.1 Input Hypothesis Overview**
The Input Hypothesis suggests that language acquisition occurs when learners are exposed to language input that is slightly beyond their current level of understanding. It emphasizes the importance of comprehensible input in the language learning process, proposing that this input helps learners acquire language subconsciously. The Input Hypothesis, by Stephen Krashen, changes how we think about learning language. Krashen once stated the "Input Hypothesis" in this way: "Only a second language learner receives the comprehensible input, which means the second language input is slightly higher than the learner's current language level, and focuses on the comprehension of 'meaning' or 'message' rather than on 'form', acquisition can be produced and the second language can be learned effectively". Krashen called this comprehensible input "i+1". "i" referred to the learner's current language level and "1" referred to language materials slightly higher than the learner's current language level (Xia, 2019). This approach focuses a lot on learning a language without realizing it, which is different from the usual way of being taught through rules and exercises. This theory says that people who are learning a language do the best when they are exposed to material that is slightly harder than what they already know. This helps them learn more while still being able to understand some of it (Krashen, 2023).

The Input Hypothesis suggests that learning a language naturally is similar to how children learn their first language by hearing and understanding it. By giving learners exposure to slightly more difficult material, they are encouraged to naturally absorb and learn the language without even realizing it. This change in focus from talking to listening agrees with the belief that understanding what is said comes before speaking well. Unlike the traditional way of thinking about how people become fluent in a foreign language, First, we learn structures. Then, we practice using them in communication. The input hypothesis says that we focus on understanding meaning first, and then we will naturally learn the structure. This matches well with the idea of how we teach languages using communication. Language production happens when people hear and learn language from others. Listening and reading are more important than speaking and writing when it comes to learning a language (Liu, 2023). Speaking is something that comes naturally when we have listened to a lot of things. Listening is an important activity that helps us learn and develop our writing and even speaking skills. Writing comes naturally after reading a lot. Reading is an important part of learning how to write

The idea of the Affective Filter, which is part of Krashen's theory, talks about how motivation, anxiety, and other feelings influence language learning. When someone is feeling motivated, relaxed, and confident, it becomes easier for them to learn a language (Krashen, 2023).

The Input Hypothesis has had a big impact on language education. It has made educators think about different ways of teaching and focus on understanding the language. The idea that students learn better when they are given challenging but understandable things to learn has caused changes in how teachers teach and what materials are used for language learning. Although people have criticized and debated the Input Hypothesis, it still plays a major role in discussions about the best methods for teaching and learning languages. The influence of this continues to shape how we learn languages and inspire new ways to help people learn languages.

**1.2 Application in Language Learning**
Several studies have explored the application of Input Hypothesis in language learning contexts. Researchers have examined its effectiveness in promoting listening and reading comprehension. It found that under the guidance of Input Hypothesis Theory, it had obviously enhanced students' performance in English reading ability. Previous findings revealed that after the reading program of 10 weeks, Polytechnic students English reading ability was significantly improved (Xia, 2019).
Research conducted by VanPatten & Williams (2007), Patten and Benati (2010), and Shimanskaya (2018) has underscored the importance of language input.
It is essential for language learners to develop their L2 proficiency and is seen as a fundamental source of cognitive depiction of the language. Consequently, the importance and purpose of language input have been established. A number of language learning theories, such as the behaviourist, mentalist, and interactionist, Ellis (2008) supports this idea whose theories were published in 2008.

## 2. English Writing Proficiency in Polytechnic Students

The need for learners to receive input in a language is accepted. linguistic ability, does not always have the same level of importance in second language acquisition theories.
Language acquisition is the process of learning and understanding language. important factor in second language acquisition.A very important thing for learning. Previously stated, language input is seen as the primary origin of linguistic information.
The learner is vulnerable (Long, 1982). Ellis (1994) states that for SLA to take place, two conditions must be met.Learners have access to L2 input and have an understanding of how L2 linguistic features are handled.
Gass (1991, 1994, 1997) and Mackey & Gass (2015) talked about the importance of input. They believed that input is not only important for understanding, but also for creating input that can be easily understood.

According to Gass, what really matters in the language input is how well we understand it, because this is important for learning a second language. fully accounted for by the traditional view of the input as raw material for L2 acquisition. This means that when learning a second language, the language input that a learner receives is more extensive than what is typically considered as the raw material for language learning.

To help to reduce the amount of understandable information that is received or understood, the effectiveness of language input depends on making it easy to understand. this can only happen when the learners understand and remember the information they receive and think about how it relates to what they already know.

Change the word "intake" to "input". L2 learners come across various language inputs that may not always be within their language abilities. Some learners may struggle with certain grammatical structures or pronunciations while finding others easier to master. Similarly, some learners may have difficulty understanding or producing certain sounds or tones in a second language. These differences in perception and ability are influenced by various factors such as individual learning styles, language background, and exposure to different languages.

Krashen's theory of second language acquisition suggests that consistent access and exposure to language input should only be beneficial when it is understood by second language learners who have a limited ability to process emotions.

The Input Hypothesis strongly suggests that second language acquisition is only possible when learners are exposed to certain aspects of the second language that they can fully understand, regardless of whether the language has been adapted to their level or higher, and has been enriched with a variety of language forms. Krashen (1989) suggested that comprehensible language input has both a positive and negative impact on second language acquisition. Krashen (1989) suggested that second language learners may experience second language acquisition when they are exposed to a type of language input that is more understandable than their language abilities.

Other people, because of different things like things they already know and have experienced before, how often some things happen compared to others, and how much attention learners give to new things they are learning.

In her theory about learning a second language, Gass (1997) says that the quality of the language input is responsible for how well someone learns the language.

This text is about a concept called "comprehension. " But instead of focusing on that term, it talks about how well someone knows a language.

**2.1 Challenges Faced by Polytechnic Students**

In China, polytechnic directional system is three years or five years, which are sent up for cultivating talents with strong practical ability. If learn skills used correctly when in school, students can mount guard hired directly after graduation. This kind of colleges and universities keeps the principle of "enough is fine, practical is the direction", the purpose of students' learning is very clear that is to learn something enough to work. In On strengthening the vocational education personnel training work opinion, the ministry of education pointed out that higher vocational education to cultivate higher technology applied special talents tat support the party's basic line, adapt to the production, construction, management and service first line need (Li, 2016).

Polytechnic students often encounter challenges in English writing, including limited time, insufficient language exposure, and the need to balance multiple subjects. These challenges can hinder their writing proficiency.

Polytechnic students who are learning English as their second language face many different problems when it comes to writing in English. These challenges are about being good at language, understanding difficult words, grammar mistakes, cultural and academic differences, having too much work to do, getting specific writing tasks, and not having enough help with writing. Research in this field focuses on finding the difficulties that polytechnic students face in writing in English and understanding the effects of these difficulties. The goal is also to create plans and resources that can assist students in improving their English writing abilities. These skills are crucial for their academic achievements and their future professional paths in a world that is connected globally.

The author's experience in teaching practice has shown that the teacher's approach is still traditional, and the teacher simply follows the traditional teaching style. A few students find it difficult to provide a response. What are the implications of the text? Students are still having difficulty comprehending the primary concept in the writings. In addition, students have a limited range of words, leading to difficulty comprehending the texts (Dang, 2023). If a learner is not familiar with the features of the text, it could lead to an issue. The specific text contains the necessary or existing information. Therefore, it is evident that a considerable amount of students are not cognizant of the difficulties associated with expressing their thoughts, and they find writing to be uninteresting (Weigle, 2016). The act of writing is not only an individual product, but also a social and cultural context, which is in line with what was stated (Maldin, 2022).

**2.2 The Importance of English Writing**

Proficiency in English writing is essential for polytechnic students, as it impacts their ability to convey ideas, complete assignments, and excel in their academic and professional endeavors. English has always been an important part of foreign language learning, especially in polytechnic. It is known to all that although input is very important in foreign language learning, it is not easy to get access to various kinds of input information. Consequently, English writing has become a significant source of information (Fei, 2023). Given that writing is one of the most straightforward ways to gain access to information, and the many benefits it can bring, it is essential for educators to make the most of their English writing lessons. This research seeks to investigate an English writing teaching approach that takes into account the positive aspects of the input hypothesis, as well as the idea of China's new curriculum reform.

## 3. The Role of Comprehensible Input in Writing

Comprehensible input, as emphasized by the Input Hypothesis, can foster vocabulary acquisition, grammar development, and idea organization. When students receive input slightly above their current writing proficiency, it may lead to improvements in their writing skills. The Input Hypothesis, a fundamental idea in language acquisition theories, has a major influence on the growth of writing abilities. This input acts as an impetus for a range of writing activities, such as expanding one's lexicon, refining grammar, and structuring thoughts in a logical way. When students are presented with material that is slightly more understandable than their current writing skills, it serves as a foundation for improving their writing proficiency.

According to the former study, it found that learners acquired second language by receiving comprehensible input and learners' language acquisition faculty. But Swain was critical to the Input Hypothesis that the interactions and comprehensible input were insufficient for successful SLA. "Conversational exchanges... are not themselves the source of acquisition derived from comprehensible input. Rather they are the source of acquisition derived from comprehensible output: output extends the linguistic repertoire of the learner as he or she attempts to create precisely and appropriately the meaning desired (1985:252)", Swain stated.

Gaining a strong command of vocabulary is greatly enhanced by input that is easy to understand. When students come across unfamiliar words and phrases in their environment, they not only understand their definitions but also become adept at utilizing them in their writing. This exposure enhances the range of words used, allowing students to express their ideas with greater accuracy and clarity in their writing. Immersion of the language is the ticket for L2 learners to clutch the point of intrinsic logic of target language. Swain stated that although immersion students who learned French over a period of eight years, they were not proficient in target language. French proficiency of the immersion students was more advanced than students who took 20, 30 minutes a day for foreign second language (FSL) learning (Li & Lu, 2015).

The comprehension of input has a profound impact on the development of grammar. Those who are presented with input that is well-organized and grammatically accurate are more likely to take these patterns into account and incorporate them into their writing. This exposure aids in minimizing grammatical mistakes and enhancing the organization of sentences in their compositions.

Furthermore, understandable input assists in structuring thoughts. When students interact with texts that provide logical and coherent information, they are more likely to emulate their own writing style. By mastering the art of structuring their thoughts, they can create compositions that are more lucid and coherent.

The Input Hypothesis proposes that supplying students with material that is slightly more difficult than their current writing ability can be especially advantageous. This approach motivates students to expand their skills and strive for greater writing proficiency. It compels them to venture beyond their comfort zones and relentlessly pursue enhancement in their writing abilities. Teachers should try to get students interested in learning by taking into consideration their circumstances. The teacher needs to understand the general characteristics of the polytechnic students, without focusing too much on each student's uniqueness. Polytechnic students' intelligence and thinking skills have reached a consistent level. Teaching reading in the polytechnic is a crucial part of language teaching, along with listening, speaking, and writing (Fei, 2023). This part can help improve the teaching of the other three aspects if done properly and based on science. If teachers discover new ways to make students interested in learning, they can help students become better at English during their three years in the polytechnic. Many VSHS students find it challenging to learn to write, particularly in EFL situations with minimal exposure to authentic English usage. Therefore, inadequate English writing skills are typically the leading cause of employment failure following graduation (Siregar et al., 2022). As a result, having much exposure to the material is essential for honing one's writing abilities. It broadens the lexicon, refines syntax, and bolsters the structure of thoughts. Educators can motivate students to become more proficient and self-assured writers by providing them with input that surpasses their current abilities, thus fostering their growth and excellence in writing.

## 4. The Effect of Input Hypothesis on Polytechnic Students

**4.1 Empirical Studies**
There is a dearth of empirical studies exploring the direct impact of Input Hypothesis on the English writing proficiency of polytechnic students. Related research in this area is a better understand the practical applications of this theory. It is widely acknowledged that having strong verbal communication skills is essential. Li's (2011) research revealed that positive vocabularies and prefabricated language chunks have the greatest impact on discourse proficiency. The second highest ranking is given to writing, applied grammatical knowledge, and self-mending ability.

Uggen employed a variety of techniques to analyze the output-input and output sequences of second language learners, mirroring Izumi and Bigelow's research(Z. Li & Lu, 2015). The results of the test did not suggest that output had an effect on learners' focus on grammar in the following text, however, the qualitative recall data showed how output affects their recognition of vocabulary and/or their knowledge of grammar rules. Literature and scholars Fang and Xia have conducted empirical research on the possibility of output driven hypothesis in oral, writing and interpretation courses(Z. Li & Lu, 2015). It was later refined to "output driven, input-enabled" theory, emphasizing the importance of input and output in college English. She and her team made a preliminary attempt to develop a new theory of foreign language teaching in mainland China. Research shows that many higher vocational students are not interested in learning English. The survey found that 35. 43% of students want to get a diploma to get ready for exams, 58. 60% need a certificate for a foreign language to increase their chances of success in their future job, and only 2. 37% are excited to learn more about English because they find it interesting. English classes are not very effective because of two main reasons: social pressure and the fact that almost all students (more than 97%) do not take the initiative to learn(L. Li, 2016). Higher vocational students have more choices in what they use to learn, so their English skills can be really different. It is clear that Shanghai students are better at speaking English than students from other countries. So, the basic English skills of the student can be divided into different levels. In class, some students might have difficulty understanding the stuff, while others might learn more. It is sad that students in high school or vocational education often ignore the basic literacy classes and may even not pay attention to the basic academic training, which leads to a lack of understanding of important theories and thinking skills. Therefore, language training this relatively boring project is interesting and ability is not enough(L. Li, 2016). The input hypothesis is about how learners pay attention to the words they use. This special attention helps students realize when their words don't say exactly what they meant. They can learn this from themselves or from others. An internal source is when a writer thinks about how readers will react to their writing. External sources can be other examples of the second language or when other people review and help point out mistakes. The Peer Feedback Intervention was made to help students find errors in their classmates' essays, but it's possible that they also learned to look at their own writing more critically, as shown by how they interacted with the reviewers. Importantly, students only gave feedback on one partner's essay at a time(Zhang & McEneaney, 2020). This allowed them to focus and understand the criteria better, as well as provide more specific and detailed feedback on personalized evaluations. This finding agrees with previous research that shows the advantages of using the input hypothesis for second language writing. The input hypothesis recommends flooding the classroom with comprehensible input in the form of natural language used for real communication (Higgs, 1985).

**4.2 Potential Benefits**
Although limited, anecdotal evidence and indirect research suggest that applying the principles of the Input Hypothesis to writing instruction may lead to improved writing skills in polytechnic students. Further research is required to validate these claims. In class, lessons can sometimes be unpredictable on the daily basis. Teachers need to be prepared to listen to what learners say, not just how they say it, and communicate with them in a natural way. They also need to use more management skills compared to a traditional classroom where the teacher has more control. Furthermore, people who do not speak English as their first language may need to have a better understanding of the language or rather, a different combination of language skills in order to communicate fluently and discuss a wider variety of topics about how the language is used than they are used to. Teachers might have to uncover hidden ideas about how to teach language (Thompson, 1996). The theory can be changed to fit different teaching situations and materials. Teachers can create writing assignments and choose reading materials that are suitable for their students' current skill levels. The theory of adapting writing tasks and reading materials is a new area of research that provides chances for more studying, testing, and creating new ideas in language education.The Input Hypothesis theory is usually used for language listening and reading. However, it could also help with writing and language development. This means it could be a good approach for teachers and researchers looking for ways to improve writing skills in students, even in polytechnic students.

## 5. Recommendations and Future Directions

**5.1 Incorporating Input Hypothesis in Writing Instruction**
Educators and institutions should consider implementing strategies that align with Input Hypothesis, such as providing comprehensible input materials and scaffolded writing tasks to enhance polytechnic students' writing proficiency. This means picking texts and subjects that polytechnic students enjoy and find useful. Using images and animations helps students comprehend and retain important concepts. Pictures and moving pictures can be images or animations that show or describe things.
Polytechnic students have different abilities when it comes to language. Therefore, it is crucial to customize teaching methods in order to meet the specific needs of each individual. Providing students with different materials to choose from at different difficulty levels can help them choose content that matches their abilities. This helps people become more independent and motivated.
Make ways for students to get feedback on their writing that helps them understand and improve it. Feedback should not only be about finding errors, but also about identifying moments when comprehending the text improved the quality of the writing.

Consider using videos, audio recordings, and interactive online content to make learning a language more enjoyable and interesting. These can improve what is written and make it simpler for students to understand.

**5.2 Future Research**

Future research should focus on empirical studies that directly investigate the impact of Input Hypothesis on English writing proficiency in polytechnic students. Longitudinal studies and controlled experiments can provide more conclusive insights into the benefits of this approach. Studying how the Input Hypothesis affects the writing skills of polytechnic students in English is a promising path to explore through actual research. Teachers and schools can change and make their teaching better as they learn from research. The benefits include teaching that is tailored to individual needs, improved educational outcomes, and graduates who are better prepared for finding jobs.

People who make rules about education should think about the findings from these studies when they create plans for what students learn and how teachers teach. If the information shows that using Input Hypothesis-based methods helps, it might be a good idea to use them in schools to help polytechnic students get better at writing.

More research can help make specialized training programs for teachers. Teachers can be given strategies to use the Input Hypothesis effectively, which will help them teach English writing better in polytechnic schools.

It is really important to be good at English to do well in school and at work. The findings of a study can directly affect how well-prepared polytechnic students are for their future careers. People who are really good at writing are more ready to start working and can make a bigger difference in many different areas.

Being able to speak English well is really important for talking to people from your own country and from around the world, especially because our world is getting more connected all the time. This research could have a big impact on how English is taught and how people communicate internationally if the teaching method based on the Input Hypothesis can help improve writing skills in English.

To put it simply, it is very important for researchers to study how the Input Hypothesis affects the English writing skills of polytechnic students. This research is important for academic reasons, but it also has practical value for teachers, organizations, lawmakers, and students themselves. By doing long-term research and careful trials, we can create better ways of teaching that might improve the future success of polytechnic graduates.

## Conclusion

In short, the Input Hypothesis theory, which focuses on language listening and reading, can improve the English writing skills of polytechnic students. However, there is not enough research on this topic, and more studies are necessary to prove its practical use in teaching writing. Learning to write well in English is very important for success in school and work. Discovering and using the Input Hypothesis can help polytechnic students improve their language skills. This in-depth review of literature helps to establish a basis for future research and instructional development focused on improving the writing abilities of polytechnic students. It is very important for teachers, schools, and researchers to keep studying, changing, and improving these methods to help polytechnic students do well in their chosen areas. Simply put, the Input Hypothesis theory suggests that improving our language listening and reading skills can help polytechnic students become better at writing in English. However, it is clear that there is not enough research in this specific area. Therefore, more studies are needed to prove how it can be applied practically in teaching writing. Being good at writing in English is definitely a very important skill, you need it to succeed in school and in your career.

For students studying technical or vocational subjects at polytechnic, it's really important to be able to write in English and explain their ideas and knowledge. The Input Hypothesis offers a way for teachers and schools to help students improve their language abilities. Using this application can help you write better and also improve your language skills overall.

This review of literature provides a strong basis for future research and instructional development. It aims to improve the writing skills of polytechnic students. The importance of this mission cannot be emphasized enough. It is very important for teachers, schools, and researchers to keep learning, changing, and improving their teaching methods. Polytechnic students can only be successful and make a positive impact in their industries if they continue to work hard and improve themselves. The progress of teaching languages is connected to the success and advancement of polytechnic education. It is important for everyone involved to support this.